%% file: main.tex
\documentclass[11pt]{article}

\usepackage[margin=1in]{geometry}
\usepackage{amsmath,amssymb}
\usepackage{booktabs}
\usepackage{multirow}
\usepackage{graphicx}
\usepackage{caption}
\usepackage{microtype}
\usepackage[hidelinks]{hyperref}
\usepackage{natbib}
\usepackage{authblk}
\makeatletter
\renewcommand\AB@affilsepx{, \protect\Affilfont}
\makeatother
\usepackage[normalem]{ulem}
\useunder{\uline}{\ul}{}

\title{\textbf{A Mechanistic Study of Automated Failure Attribution in LLM-based Agentic Systems}}
\author[1]{Wendy Zheng}
\author[2]{Liang Wu}
\author[1]{Yinhan He}
\author[2]{Mayank Darbari}
\author[1]{Jundong Li}
\author[2]{Liangjie Hong}

\affil[1]{University of Virginia} 
\affil[2]{Nokia}
\date{}

\begin{document}
\maketitle

\begin{abstract}
\input{sections/0-abstract}
\end{abstract}

\section{Introduction}\label{sec:introduction}
\input{sections/1-introduction}

\section{Background and Preliminaries}
\input{sections/2-preliminary}

\section{Method}
\input{sections/3-method}

\section{Experiments}
\input{sections/4-experiments}

\section{Related Works}
\input{sections/5-related_works}

\section{Conclusion}
\input{sections/6-conclusion}

\bibliographystyle{iclr2027_conference}
\bibliography{iclr2027_conference}

\end{document}

%% file: sections/0-abstract.tex
Agentic systems have demonstrated strong capabilities on complex tasks by enabling multiple interacting large language model (LLM)-based agents to collaborate and execute subtasks in parallel. However, this collaborative structure introduces new challenges for reliably evaluating and diagnosing system failures, particularly failure attribution: tracing a system's execution to identify the agent step responsible for an incorrect final outcome. Because manually performing this task is time- and labor-intensive, recent work has turned to LLMs as judges to automate the process, referred to as automated failure attribution (AFA). These approaches assume that a sufficiently capable model can infer the failing step directly from an interaction transcript. In practice, however, direct prompting often performs poorly, which motivates follow-up work that explores various ways to improve AFA performance, including alternative prompting strategies and inference-time interventions. In this paper, we examine a more fundamental question: whether this weak behavioral performance reflects a genuine capability limitation. Using linear probes applied to internal activations, we uncover a significant \emph{knows-but-won't-say gap:} models encode sufficient information to attribute a failure correctly but fail to explicitly output the responsible step. Building on this finding, we propose \emph{grafting}, a simple manual intervention that exploits knowledge already encoded in the model to improve AFA performance without requiring ground-truth labels at inference time. Extensive evaluations across multiple LLMs and benchmark settings demonstrate that grafting consistently outperforms direct-prompting approaches, underscoring the potential of leveraging latent model knowledge for more reliable failure attribution. Our code is available at~\url{https://github.com/wendyzheng22/mi4afa}.

%% file: sections/1-introduction.tex
Agentic systems extend the capabilities of large language models (LLMs) by decomposing complex tasks and distributing them across multiple specialized agents (e.g., planners, retrievers, coders, and verifiers), that collaborate and execute subtasks in parallel~\citep{acharya2025agentic,bandi2025rise,wu2023autogen}. This collaborative paradigm enables agentic systems to tackle problems that would exceed the capacity of a single model working alone. However, as the deployment of such systems expands into high-stakes settings, reliable evaluation methods become essential~\citep{tran2025multi,gridach2025agentic}. One key evaluation task is failure attribution: identifying the earliest step in a system's trajectory that caused the final output to fail. Performing this task manually is expensive, since agentic systems can involve many interacting agents, and the space of possible interactions grows exponentially with system size. This challenge motivates Automated Failure Attribution (AFA)~\citep{zhang2025agent}, which employs an LLM as a judge to identify the failing step directly from a system's interaction transcript. However, as demonstrated by the Who\&When benchmark~\citep{zhang2025agent}, even large, capable LLMs perform poorly on this task.

This weak performance has spurred a line of follow-up work that seeks to improve AFA, such as refined prompting strategies~\citep{west2025abduct,sun2026scope} or inference-time interventions~\citep{ma2026dover}. These approaches share a common underlying assumption: the capability for failure attribution already exists within the model and simply needs to be elicited more effectively. However, the persistently poor performance of direct prompting raises a more fundamental question: do LLMs actually possess the capability required for AFA? To answer this, we look beyond the model's generated output and instead examine the information encoded in its internal representations using simple linear probes~\cite{alain2016understanding}. Surprisingly, these probes recover substantially more information about the correct failing step than the model's generated answer. We refer to this discrepancy as the \emph{knows-but-won't-say gap}: LLMs encode considerably more AFA-relevant information internally than they are able to express behaviorally. This finding indicates that poor AFA performance does not necessarily reflect an absence of underlying capability; rather, the relevant information may already be present within the model but not effectively accessed during generation.

This observation then raises a second question: if models already encode the information needed for AFA, can this latent capability be translated into better behavioral performance? To investigate this question, we propose \emph{grafting}, a simple manual intervention that connects internal components encoding information about the failing step to the components involved in generating the model's output. Rather than teaching the model new failure-attribution knowledge, grafting improves how existing information is routed during generation, enabling the model to more effectively express knowledge it already possesses. In this way, grafting converts latent, decodable AFA information into improved behavioral predictions without requiring ground-truth labels at inference time. 

We evaluate direct prompting, linear probing, and grafting across multiple open-weight LLMs and benchmark settings, and find a consistent progression: direct prompting provides a weak behavioral baseline; linear probes reveal that substantially more information about the correct failing step is encoded in model activations; and grafting converts a significant portion of this latent information into improved behavioral predictions. Together, these findings suggest that the limitations of AFA stem not solely from whether models contain the relevant information, but also from whether that information can be effectively accessed. Our contributions are summarized as follows:

\begin{itemize}
\item \textbf{Knowledge--behavior gap.} We show that linear probes consistently outperform direct prompting across model--dataset combinations, revealing a substantial gap between the AFA-relevant information encoded in model activations and the information expressed through behavioral predictions. This provides evidence that LLMs possess considerably more AFA-relevant capability than their direct responses reflect.
\item \textbf{Grafting.} We introduce \emph{grafting}, a simple manual intervention that bridges components that encode information about the failing step to those involved in output generation. Grafting consistently improves AFA performance over direct prompting without requiring ground-truth labels at inference time.

% \item \textbf{Mechanistic analysis.} We identify the internal components involved in grafting and, through ablation studies, show that the identified attention heads are functionally important for its success, rather than the observed improvements arising from generic attention perturbations.
\end{itemize}

%% file: sections/2-preliminary.tex
In this section, we formally define the Automated Failure Attribution (AFA) task and establish the performance of direct prompting as a baseline. These results motivate the central question of this work: whether the poor behavioral performance of LLMs on AFA reflects a genuine lack of capability, or merely a failure to express information already encoded within the model.

\subsection{Automated Failure Attribution}
Consider an agentic system composed of one or more interacting agents that is given a task but produces an incorrect final answer. Let $H = (h_1,h_2,\ldots,h_n)$ denote the resulting interaction transcript, where each $h_i$ corresponds to the output of one agent at step $i$. Each transcript is annotated with the earliest step $e \in \{1,\ldots,n\}$ whose error ultimately causes the system to fail, which referred to as the \emph{mistake step} (equivalently, the \emph{failure step}). Given the transcript $H$, the goal of AFA is to predict the failing step $\hat e \in \{1,\ldots,n\}$. An AFA method succeeds when $\hat e = e$, correctly identifying the earliest mistake responsible for the incorrect final output.

Following~\citet{zhang2025agent}, we report two evaluation metrics throughout the paper: (1) \textbf{Step accuracy} measures the probability of correctly identifying the failing step, $P(\hat e = e)$, and (2) \textbf{Agent accuracy} measures whether the agent at the predicted mistake step $\hat e$ matches the ground-truth failing agent, regardless of which occurrence of that agent is selected. Because a single agent may speak multiple times over the course of a transcript, a prediction can identify the wrong step while still attributing failure to the correct agent. Agent accuracy is therefore a strictly coarser metric than step accuracy, and the two do not necessarily move together.

\subsection{Direct Prompting}
We first evaluate direct prompting to characterize the behavioral baseline of LLMs on the AFA task, using the Who\&When benchmark~\citep{zhang2025agent}. This benchmark consists of two subsets: the \textbf{Algorithm-generated (Algo.)} subset, containing relatively short execution traces produced automatically, and the \textbf{Hand-crafted (Craft.)} subset, containing substantially longer and more complex agent trajectories designed by human annotators.

Table~\ref{tab:orig_prompting_results} summarizes the performance of several open-weight and closed-source LLMs under direct prompting. The closed-source LLMs are accessed via Cursor Python SDK~\citep{cursor_python_sdk}. Two observations are notable. First, consistent with prior findings, every model achieves less desirable step accuracy on both subsets, confirming that reliably identifying the failing step remains difficult regardless of model family. Second, performance does not scale consistently with model size or capability: within the Qwen family, the 1.5B model underperforms the 0.5B variant on step accuracy for Algo., and the closed-source models only modestly outperform substantially smaller open-weight models on Craft. This absence of a clear capacity trend suggests that increasing model size alone is insufficient to close the gap in behavioral performance on AFA, motivating the investigation, in the following sections, of what the model encodes internally rather than what direct prompting can elicit from it.

\begin{table}
\centering
\caption{Direct Prompting Results on the Who\&When benchmark}
\label{tab:orig_prompting_results}
\begin{tabular}{ccccc}
\toprule
              & \multicolumn{2}{c}{Algorithm Generated} & \multicolumn{2}{c}{Hand Crafted} \\ \midrule
              & Agent Acc.          & Step Acc.         & Agent Acc.      & Step Acc.      \\ \midrule
Qwen 0.5B     & 41.27               & 22.22             & 39.29           & 7.14           \\
Qwen 1.5B     & 33.33               & 7.94              & 46.43           & 10.71          \\
Qwen 7B       & 30.16               & 12.70             & 46.43           & 0.00           \\
Llama 1B      & 23.81               & 15.87             & 17.86           & 3.57           \\
Llama 3B      & 30.16               & 17.46             & 25.00           & 7.14           \\
Gemma 1B      & 28.57               & 9.52              & 32.14           & 3.57           \\
Gemma 4B      & 26.98               & 7.94              & 60.71           & 7.14           \\
GPT-5.6 Sol   & 61.90               & 50.79             & 57.14           & 25.00          \\
GPT-5.6 Terra & 65.08               & 52.38             & 67.86           & 35.71          \\
Sonnet 5      & 57.14               & 42.86             & 67.86           & 28.57          \\
Opus 4.8      & 61.90               & 46.03             & 78.57           & 39.29          \\
Opus 5        & 66.67               & 50.79             & 82.14           & 35.71          \\ \bottomrule
\end{tabular}
\end{table}

%% file: sections/3-method.tex
In this section, we introduce two methods to address the questions posed in Section~\ref{sec:introduction}: (1) \emph{probing}, which determines whether LLMs internally encode sufficient information to identify the failing step, and (2) \emph{grafting}, which investigates whether this latent information can be exposed to improve the model’s behavioral performance.

\subsection{Probing: Is the failing step decodable from model activations?}
Poor performance under direct prompting does not necessarily mean that an LLM lacks the capability required for AFA. The relevant information may instead be encoded in its hidden representations without being reflected in its output distribution. To distinguish between these possibilities, we train linear probes to test whether the failing step can be linearly decoded from intermediate activations. Our probing pipeline has three stages: activation extraction, probe training, and prediction.

\paragraph{Activation extraction.}
For each transcript, we perform a single forward pass through the frozen LLM and extract hidden representations from every transformer layer at two sites: the post-attention residual stream and the post-MLP residual stream. For a model with $L$ layers, this gives $2L$ candidate representation sites. Because AFA is a step-level prediction problem, each step is represented by the hidden state $x$ at the final token of that step.

\paragraph{Probe training.}
We evaluate each of the $2L$ representation sites independently. For a fixed site, we pool the hidden states $x$ from every step in every training transcript into a single dataset. Each hidden state receives a binary label indicating whether its corresponding step is the ground-truth failing step:
\[
y =
\begin{cases}
1,&\text{if the step is the ground-truth failing step},\\
0,&\text{otherwise}.
\end{cases}
\]

We train a linear logistic probe
\begin{equation}
    s(x)=w^\top \hat{x}+b,
\end{equation}
where 
\begin{equation}
    \hat{x}=\frac{x-\mu}{\sigma},
\end{equation}
and $\mu$ and $\sigma$ are computed only on the training split. The probe minimizes

\[
\mathcal{L}(w,b)
=
\mathbb{E}
\left[
\mathrm{BCE}(s(\hat{x}),y;\pi)
\right]
+
\lambda\|w\|_2^2,
\]
where
\[
\pi=\frac{N-N_+}{N_+}
\]
is the positive-class weight used to compensate for the severe class imbalance created by each transcript having exactly one labeled failure step.

\paragraph{Prediction.}
At evaluation, the trained probe independently scores every step in a transcript using the corresponding hidden representation. The predicted failing step is the highest-scoring step: $\hat e = \arg\max_k s(x_k)$. Evaluating probes at every residual location allows us to identify where failure-attribution information is most linearly accessible within the model.

\subsection{Grafting: Making the model say what it locates}
Linear probing reveals what information is encoded in the model but leaves its behavior unchanged. The goal now is to improve the model's own predictions. Our approach is motivated by a simple hypothesis about the functional organization of transformer attention heads: some heads specialize in locating the failing step (locator heads), while others specialize in writing the model's predicted answer into the output distribution (writer heads). If these two groups are only weakly coupled, the model may internally identify the correct failing step without effectively communicating this information to the output. Grafting aims to bridge this gap using a three step process: (1) identify locator heads, (2) identify writer heads, and (3) connect them through an inference-time intervention.

\paragraph{Step 1: Identify locator heads}
We define locator heads as attention heads whose attention patterns reliably identify the failing step. Intuitively, if a head consistently places most of its attention on the ground-truth mistake step when producing the final answer, then that head provides a useful signal about where the failure occurred. Thus, for each head $(\ell,h)$, we compute the attention distribution at the last token position:

\begin{equation}
    z^{(\ell,h)} = \mathrm{softmax}\!\left( \frac{q^{(\ell,h)}_{\text{last}}K^{(\ell,h)}}{\sqrt{d_h}} \right),
\end{equation}

where $d_h$ represents the head dimension. Summing the attention mass over each step's token span gives the total weight assigned to step $s$:
\begin{equation}
\alpha^{(\ell,h)}_s \;=\; \sum_{t=\tau_t}^{\tau_{t+1}-1} z_{t}^{(l,h)},
\qquad s = 1,\dots,n,
\label{eq:turnattn}
\end{equation}
where $\tau_s$indexes the token boundaries of step $s$. Each head casts a vote for the step receiving the largest attention mass,

\[
\arg\max_s \alpha^{(\ell,h)}_s.
\]

A head's \emph{locator score} is the proportion of training examples on which its vote matches the ground-truth failing step. We rank heads by this score and retain at most five heads from the top 10\% of the ranking, excluding heads in the final transformer layer (a final-layer head has no downstream computation left to route its signal into, so it cannot itself act as a useful locator for the graft).

\paragraph{Step 2: Identify writer heads}
Whereas locator heads determine \emph{which} step is important, writer heads determine \emph{what} the model ultimately outputs. We therefore identify these using direct logit attribution~\citep{logitlens2020}, which measures a component's direct contribution to a particular output logit. Each attention and MLP layer's output is added into the residual stream, so the final residual stream $\text{resid}^{(L)}$ is a linear sum of every component's contribution. Projecting a component's output through the model's final normalization scale $g$ and the unembedding matrix $W_U$ gives that component's direct, linear contribution to any vocabulary logit $v$: 
\begin{equation}
\mathrm{DLA}^{(\ell,h)}(v) \;=\; \frac{\big(W_O^{(\ell,h)} z^{(\ell,h)}_{\text{last}}\big)^\top
                                  r(v, C)}{\mathrm{rms}\big(\text{resid}^{(L)}_{\text{last}}\big)}.
\label{eq:dla}
\end{equation}
where $z^{(\ell,h)}_{\text{last}}$ is the output of an attention head at the last token (before the output projection $W_O$), and $r(v, C)$ is a contrastive readout direction defined as
\begin{equation}
    r(v, C)  = g \odot \Big( W_U[v] - \tfrac{1}{|C|}\textstyle\sum_{c \in C} W_U[c] \Big),
    \label{eq:readout}
\end{equation}
where $C$ is the set of candidate step-index tokens for the transcript. Subtracting the mean unembedding vector over the candidate tokens removes directions shared across candidates and emphasizes the dimensions that distinguish the target token from the other possible step indices. Writer heads are scored by their DLA toward the step index token that the model emits on the clean forward pass, instead of the ground truth mistake step. We intentionally measure contribution toward the model's own predicted answer because the objective is to identify the components currently responsible for expressing the model's internal belief, irrespective of whether that belief is correct. Writer heads are identified on the training set, keeping up to five heads from the top 10\%, excluding heads in the first layer (a first-layer head has not yet had access to enough contextual computation to serve as a meaningful writer).

\paragraph{Step 3: Construct the graft}
Once the locator and writer heads have been identified, grafting combines them at inference time without any supervision or ground-truth labels:
\begin{enumerate}
  \item \textbf{Read the locators.} Each selected locator head votes for its top-attended step; we take the plurality winner $\hat{t}$ as the model's internal guess. We use a vote across heads rather than relying on a single head because individual heads are not perfectly reliable.
  \item \textbf{Form the target direction.} Map $\hat{t}$ to its index token and compute $r(\hat{t})$ by Equation~\ref{eq:readout}.
  \item \textbf{Pull it back into head space.} For each writer head $i$, we compute the direction in its value space that maximally increases the target logit after the head's output projection: $v_i = W_O^{(i)\top} r(\hat{t})$.
  \item \textbf{Inject.} At the final token position, add $v_i$ to the writer's value vector before $W_O$, scaled relative to the norm of the full concatenated head output $z_{\text{last}}$ at that position: 
\begin{equation}
z_i \;\leftarrow\; z_i \;+\; \beta\,\frac{\|z_{\text{last}}\|}{\|v_i\|}\, v_i .
\label{eq:graft}
\end{equation}
Here, $\beta$ controls the strength of the intervention, while the norm ratio scales the injected direction relative to the magnitude of the model's existing head output.
\end{enumerate}

The intervention is applied in the writer heads' value space rather than directly to the residual stream. As a result, the injected signal passes through the same output projection $W_O$ that the model normally uses to write information from the head into the residual stream. Together, these steps create a bridge between the model's internal localization and its output behavior.

%% file: sections/4-experiments.tex
\begin{table}[]
\caption{Performance comparison between different prompting strategies, probing, and grafting}
\label{tab:main_experiment}
\centering
\resizebox{\textwidth}{!}{
\begin{tabular}{cccccccc}
\toprule
                         &                            &            & \multicolumn{3}{c}{Prompting}               & Probing        & Grafting       \\ \midrule
                         &                            &            & All-at-Once & Step-by-step & Sequential     &                &                \\ \midrule
\multirow{14}{*}{Algo.}  & \multirow{2}{*}{Qwen 0.5B} & Agent Acc. & 34.92       & 41.27        & 47.62          & {\ul 53.97}    & \textbf{61.90} \\
                         &                            & Step Acc.  & 23.81       & 22.22        & 14.29          & {\ul 36.51}    & \textbf{38.10} \\ \cmidrule{2-8} 
                         & \multirow{2}{*}{Qwen 1.5B} & Agent Acc. & 30.16       & 33.33        & {\ul 50.79}    & \textbf{57.14} & \textbf{57.14} \\
                         &                            & Step Acc.  & 20.63       & 7.94         & 6.35           & \textbf{44.44} & {\ul 33.33}    \\ \cmidrule{2-8} 
                         & \multirow{2}{*}{Qwen 7B}   & Agent Acc. & 28.57       & 30.16        & 39.68          & \textbf{61.90} & {\ul 49.21}    \\
                         &                            & Step Acc.  & 11.11       & 12.70        & 14.29          & \textbf{47.62} & {\ul 30.16}    \\ \cmidrule{2-8} 
                         & \multirow{2}{*}{Llama 1B}  & Agent Acc. & 25.40       & 23.81        & 30.16          & {\ul 53.97}    & \textbf{60.32} \\
                         &                            & Step Acc.  & 20.63       & 15.87        & 17.46          & \textbf{36.51} & {\ul 34.92}    \\ \cmidrule{2-8} 
                         & \multirow{2}{*}{Llama 3B}  & Agent Acc. & 38.10       & 30.16        & 50.79          & \textbf{60.32} & {\ul 53.97}    \\
                         &                            & Step Acc.  & 22.22       & 17.46        & 12.70          & \textbf{46.03} & {\ul 34.92}    \\ \cmidrule{2-8} 
                         & \multirow{2}{*}{Gemma 1B}  & Agent Acc. & 28.57       & 28.57        & 23.81          & \textbf{63.49} & {\ul 39.68}    \\
                         &                            & Step Acc.  & 14.29       & 9.52         & 9.52           & \textbf{42.86} & {\ul 23.81}    \\ \cmidrule{2-8} 
                         & \multirow{2}{*}{Gemma 4B}  & Agent Acc. & 30.16       & 26.98        & {\ul 60.32}    & \textbf{61.90} & 41.27          \\
                         &                            & Step Acc.  & 11.11       & 7.94         & 11.11          & \textbf{44.44} & {\ul 22.22}    \\ \midrule
\multirow{14}{*}{Craft.} & \multirow{2}{*}{Qwen 0.5B} & Agent Acc. & 39.29       & 39.29        & 28.57          & \textbf{67.86} & {\ul 64.29}    \\
                         &                            & Step Acc.  & 10.71       & 7.14         & 7.14           & {\ul 25.00}    & \textbf{28.57} \\ \cmidrule{2-8} 
                         & \multirow{2}{*}{Qwen 1.5B} & Agent Acc. & 32.14       & 46.43        & 39.29          & \textbf{75.00} & {\ul 57.14}    \\
                         &                            & Step Acc.  & 3.57        & 10.71        & {\ul 17.86}    & \textbf{25.00} & \textbf{25.00} \\ \cmidrule{2-8} 
                         & \multirow{2}{*}{Qwen 7B}   & Agent Acc. & 28.57       & 46.43        & \textbf{64.29} & {\ul 60.71}    & 32.14          \\
                         &                            & Step Acc.  & 0.00        & 0.00         & 7.14           & \textbf{28.57} & {\ul 10.71}    \\ \cmidrule{2-8} 
                         & \multirow{2}{*}{Llama 1B}  & Agent Acc. & 21.43       & 17.86        & 14.29          & \textbf{57.14} & {\ul 35.71}    \\
                         &                            & Step Acc.  & {\ul 3.57}  & {\ul 3.57}   & 0.00           & \textbf{21.43} & \textbf{21.43} \\ \cmidrule{2-8} 
                         & \multirow{2}{*}{Llama 3B}  & Agent Acc. & 17.86       & 25.00        & {\ul 57.14}    & \textbf{75.00} & 46.43          \\
                         &                            & Step Acc.  & 0.00        & 7.14         & 7.14           & \textbf{28.57} & {\ul 25.00}    \\ \cmidrule{2-8} 
                         & \multirow{2}{*}{Gemma 1B}  & Agent Acc. & 32.14       & 32.14        & 7.14           & \textbf{60.71} & {\ul 57.14}    \\
                         &                            & Step Acc.  & 3.57        & 3.57         & 3.57           & {\ul 17.86}    & \textbf{21.43} \\ \cmidrule{2-8} 
                         & \multirow{2}{*}{Gemma 4B}  & Agent Acc. & 42.86       & 60.71        & 50.00          & \textbf{75.00} & {\ul 67.86}    \\
                         &                            & Step Acc.  & 3.57        & 7.14         & 3.57           & \textbf{21.43} & {\ul 17.86}    \\ \bottomrule
\end{tabular}
}
\end{table}

In this section, we first describe our experimental setup and then present results comparing the performance of probing and grafting to the behavioral prompting baselines.

\subsection{Experiment Setup}
We evaluate seven open-weight LLMs spanning three model families and multiple model sizes: Qwen 2.5 0.5B, 1.5B, and 7B Instruct variants~\citep{qwen2,qwen2.5,qwen2.50.5B,qwen2.51.5B,qwen2.57B}; Llama 3.2 1B and 3B variants~\citep{llama321b,llama323b}; and Gemma 3 1B and 4B variants~\citep{gemma31b,gemma34b,team2025gemma}. This selection allows us to examine AFA performance across both model scale and model family.

\noindent\textbf{Evaluation Tasks.} We adopt the Who\&When benchmark~\citep{zhang2025agent}, which contains two subsets with different trajectory lengths. The Algorithm Generation (Algo.) subset contains transcripts with an average of 9 steps, while the Hand Crafted (Craft.) subset contains substantially longer transcripts with an average of 29 steps. This variation allows us to evaluate AFA across agentic trajectories of different complexity.

\noindent\textbf{Baselines.}
We compare against three prompting-based baselines. All-at-Once provides the entire transcript to the model at once, and Step-by-Step incrementally presents each step to the model to determine whether it constitutes the failure; both strategies were introduced by ~\citet{zhang2025agent}. Sequential decomposes attribution into two stages, first identifying the responsible agent and subsequently predicting the failing step conditioned on that agent. 

\noindent\textbf{Computation Resources.} All experiments were performed on a cluster of four A100 80GB PCIe GPUs running Cuda version 13.0.

\subsection{Failure-Attribution Information is Linearly Decodable}
Table~\ref{tab:main_experiment} compares the performance of all five methods. We first examine whether the poor performance of direct prompting reflects a genuine absence of failing-step information within the model. Across the evaluated models, linear probes substantially outperform the prompting baselines on step accuracy. For Qwen 0.5B on Algo., for example, the strongest prompting baseline reaches 23.81\% step accuracy, compared with 36.51\% for probing. For Qwen 1.5B and Qwen 7B, probing reaches 44.44\% and 47.62\%, respectively, compared with 15.87\% and 14.29\% for the strongest prompting baseline. Comparable gains are observed for Llama and Gemma. 

These results indicate that failing-step information is substantially more accessible from the model's internal representations than from its generated answers. A simple linear classifier can recover information about the correct step that the model's own output does not reliably express. We refer to this discrepancy between internally encoded latent information and behavioral performance as the \emph{knows-but-won't-say} gap.

At the same time, the results reveal substantial variation in the behavioral baselines across the different configurations. No prompting strategy consistently performs best, and increasing model size does not uniformly translate into better failure attribution. This variability further suggests that the primary challenge may not simply be acquiring failure-attribution information, but effectively accessing and expressing information that is already encoded in the model.

\subsection{Grafting Exploits Latent Information for Improved Performance}
The probing results establish that LLMs encode information about the failing step, but probing itself does not provide a mechanism for using this information during generation. This motivates our next approach, grafting, which aims to leverage the model's own output mechanism to express its predicted failing step. As shown in Table~\ref{tab:main_experiment}, grafting consistently outperforms the prompting baselines and substantially narrows the gap between behavioral prediction and the information revealed by probing. For example, on the Algo. subset, grafting achieves the highest step accuracy for Qwen 0.5B (38.10\%), while also substantially outperforming all prompting methods for Qwen 1.5B (33.33\% vs. 15.87\% for Step-by-Step) and Qwen 7B (30.16\% vs. 14.29\% for Sequential). Similar improvements are observed for the Llama and Gemma models. On the Craft. subset, grafting also improves step accuracy over prompting for several models, including Qwen 0.5B (28.57\%), Gemma 1B (21.43\%), and Llama 3B (21.43\%).

Importantly, grafting does not simply reproduce the performance of the linear probe: probing remains stronger in many settings. Rather, the results show that a substantial portion of the information that is linearly decodable from the model can be converted into improved behavioral predictions through a targeted intervention on the model's internal computation. This provides evidence that the knowledge--behavior gap is not merely an observation about representational capacity, but can be actively exploited to improve AFA.

%% file: sections/5-related_works.tex
\paragraph{Automated Failure Attribution}
Automated Failure Attribution (AFA) aims to identify the agent and step responsible for a task failure in an LLM-based multi-agent system~\citep{zhang2025agent}. The Who\&When benchmark demonstrates a substantial gap between agent-level and step-level attribution\citep{zhang2025agent}. Thus, subsequent work aims to improve AFA performance through diverse strategies, including decomposing attribution into scope delineation and localization~\citep{sun2026scope}, causal counterfactual reasoning~\citep{west2025abduct}, and intervention-based verification of failure hypotheses~\citep{ma2026dover}. Other recent approaches further investigate automated localization and hierarchical attribution~\citep{zhang2026agentracer,banerjee2025did}. More recently, TraceElephant evaluates attribution under full execution observability, showing that access to complete execution traces can substantially improve attribution accuracy~\citep{chen2026seeing}. These approaches primarily seek to improve an evaluator's ability to produce correct attribution judgments from trajectory information. We instead ask whether AFA-relevant information is already encoded in the evaluator's internal representations and whether it can be routed into generation.

\paragraph{Latent Knowledge and Probing}
Prior work has shown that language models can encode information that is not reliably expressed in their outputs. Probing classifiers provide a standard method for testing whether properties are linearly recoverable from intermediate representations~\citep{alain2016understanding,belinkov2022probing}. In particular, latent knowledge can be recovered from activations even when models are prompted to produce incorrect answers~\citep{burns2022discovering}, and hidden states encode signals related to truthfulness~\citep{azaria2023internal} with approximately linear structure~\citep{marks2023geometry}. These results motivate the distinction between information that is decodable from representations and information that is actually expressed or used during generation. We apply this distinction to AFA, using linear probes to quantify attribution information encoded in intermediate representations and testing whether this information can be made behaviorally accessible during generation.

\paragraph{Activation Intervention and Representation Engineering}
A complementary literature studies interventions on internal representations to with the goal of altering model behavior. Activation engineering and representation engineering manipulate the model's intermediate representations to steer its outputs~\citep{turner2023steering,zou2023representation}, while inference-time intervention uses directions identified from activation-space classifiers to steer models toward more truthful outputs~\citep{li2023inference}. Activation patching and causal tracing provide complementary methods for localizing model components involved in specific computations through causal interventions~\citep{meng2022locating,zhang2024towards}. Grafting builds on these ideas but targets a different problem: rather than steering toward a predefined behavioral attribute, it connects representations from which a task-specific variable (i.e., the failing step)is decodable to components involved in output generation. Thus, grafting tests whether information already decodable from a model's representations can be translated into improved task behavior.

%% file: sections/6-conclusion.tex
In this paper, we study Automated Failure Attribution (AFA) from the perspective of the gap between information encoded in LLM representations and information expressed through their behavior. We show that linear probes consistently recover substantially more information about the failing step than direct prompting, revealing a \emph{knows-but-won't-say gap} in LLM-based failure attribution. Motivated by this observation, we propose \emph{grafting}, a simple inference-time intervention that connects components encoding failing-step information to those responsible for output generation, allowing the model to more effectively express its internally encoded knowledge. Experiments across multiple LLMs and benchmark settings demonstrate that grafting improves behavioral AFA performance over direct prompting, showing that latent failure-attribution information can be translated into improved predictions without requiring ground-truth labels at inference time. More broadly, these findings suggest that improving LLM-based evaluators may depend not only on scaling models or refining prompts, but on developing methods that allow models to more effectively access and express knowledge they already possess.

%% file: iclr2027_conference.bib
@article{zhang2025agent,
  title={Which agent causes task failures and when? on automated failure attribution of llm multi-agent systems},
  author={Zhang, Shaokun and Yin, Ming and Zhang, Jieyu and Liu, Jiale and Han, Zhiguang and Zhang, Jingyang and Li, Beibin and Wang, Chi and Wang, Huazheng and Chen, Yiran and others},
  journal={arXiv preprint arXiv:2505.00212},
  year={2025}
}

@article{acharya2025agentic,
  title={Agentic AI: Autonomous intelligence for complex goals—A comprehensive survey},
  author={Acharya, Deepak Bhaskar and Kuppan, Karthigeyan and Divya, B},
  journal={IEEe Access},
  volume={13},
  pages={18912--18936},
  year={2025},
  publisher={IEEE}
}

@article{bandi2025rise,
  title={The rise of agentic ai: A review of definitions, frameworks, architectures, applications, evaluation metrics, and challenges},
  author={Bandi, Ajay and Kongari, Bhavani and Naguru, Roshini and Pasnoor, Sahitya and Vilipala, Sri Vidya},
  journal={Future Internet},
  volume={17},
  number={9},
  pages={404},
  year={2025},
  publisher={MDPI}
}

@article{tran2025multi,
  title={Multi-agent collaboration mechanisms: A survey of llms},
  author={Tran, Khanh-Tung and Dao, Dung and Nguyen, Minh-Duong and Pham, Quoc-Viet and O'Sullivan, Barry and Nguyen, Hoang D},
  journal={arXiv preprint arXiv:2501.06322},
  year={2025}
}

@article{gridach2025agentic,
  title={Agentic ai for scientific discovery: A survey of progress, challenges, and future directions},
  author={Gridach, Mourad and Nanavati, Jay and Abidine, Khaldoun Zine El and Mendes, Lenon and Mack, Christina},
  journal={arXiv preprint arXiv:2503.08979},
  year={2025}
}

@article{west2025abduct,
  title={Abduct, act, predict: Scaffolding causal inference for automated failure attribution in multi-agent systems},
  author={West, Alva and Weng, Yixuan and Zhu, Minjun and Lin, Zhen and Ning, Zhiyuan and Zhang, Yue},
  journal={arXiv preprint arXiv:2509.10401},
  year={2025}
}

@article{wu2023autogen,
  title={Autogen: Enabling next-gen llm applications via multi-agent conversation},
  author={Wu, Qingyun and Bansal, Gagan and Zhang, Jieyu and Wu, Yiran and Li, Beibin and Zhu, Erkang and Jiang, Li and Zhang, Xiaoyun and Zhang, Shaokun and Liu, Jiale and others},
  journal={arXiv preprint arXiv:2308.08155},
  year={2023}
}

@inproceedings{ma2026dover,
  title={Dover: Intervention-driven auto debugging for llm multi-agent systems},
  author={Ma, Ming and Zhang, Jue and Yang, Fangkai and Kang, Yu and Lin, Qingwei and Rajmohan, Saravan and Zhang, Dongmei},
  booktitle={International Conference on Learning Representations},
  volume={2026},
  pages={28059--28087},
  year={2026}
}

@inproceedings{sun2026scope,
  title={Scope delineation before localization: A two-stage framework for enhancing failure attribution in multi-agent systems},
  author={Sun, Kai and Li, Wenqiang and Dong, Bo and Lin, Yuxin and Zhang, Jingyao and Shi, Bin},
  booktitle={Proceedings of the AAAI Conference on Artificial Intelligence},
  volume={40},
  number={39},
  pages={33108--33116},
  year={2026}
}

@article{alain2016understanding,
  title={Understanding intermediate layers using linear classifier probes},
  author={Alain, Guillaume and Bengio, Yoshua},
  journal={arXiv preprint arXiv:1610.01644},
  year={2016}
}

@misc{logitlens2020,
  author    = {nostalgebraist},
  title     = {interpreting GPT: the logit lens},
  year      = {2020},
  date      = {2020-08-30},
  url       = {https://www.lesswrong.com/posts/AcKRB8wDpdaN6v6ru/interpreting-gpt-the-logit-lens},
  organization = {LessWrong}
}

@article{meng2022locating,
  title={Locating and editing factual associations in gpt},
  author={Meng, Kevin and Bau, David and Andonian, Alex and Belinkov, Yonatan},
  journal={Advances in neural information processing systems},
  volume={35},
  pages={17359--17372},
  year={2022}
}

@inproceedings{zhang2024towards,
  title={Towards best practices of activation patching in language models: Metrics and methods},
  author={Zhang, Fred and Nanda, Neel},
  booktitle={International Conference on Learning Representations},
  volume={2024},
  pages={1651--1678},
  year={2024}
}

@article{turner2023steering,
  title={Steering language models with activation engineering},
  author={Turner, Alexander Matt and Thiergart, Lisa and Leech, Gavin and Udell, David and Vazquez, Juan J and Mini, Ulisse and MacDiarmid, Monte},
  journal={arXiv preprint arXiv:2308.10248},
  year={2023}
}

@inproceedings{zhang2026agentracer,
  title={Agentracer: Who is inducing failure in the llm agentic systems?},
  author={Zhang, Guibin and Wang, Junhao and Chen, Junjie and Zhou, Wangchunshu and Wang, Kun and Yan, Shuicheng},
  booktitle={International Conference on Learning Representations},
  volume={2026},
  pages={11377--11399},
  year={2026}
}

@article{banerjee2025did,
  title={Where did it all go wrong? a hierarchical look into multi-agent error attribution},
  author={Banerjee, Adi and Nair, Anirudh and Borogovac, Tarik},
  journal={arXiv preprint arXiv:2510.04886},
  year={2025}
}

@inproceedings{chen2026seeing,
  title={Seeing the whole elephant: A benchmark for failure attribution in llm-based multi-agent systems},
  author={Chen, Mengzhuo and Wang, Junjie and Mu, Fangwen and Wang, Yawen and Liu, Zhe and Feng, Huanxiang and Wang, Qing},
  booktitle={Proceedings of the 64th Annual Meeting of the Association for Computational Linguistics (Volume 1: Long Papers)},
  pages={19888--19905},
  year={2026}
}

@article{belinkov2022probing,
  title={Probing classifiers: Promises, shortcomings, and advances},
  author={Belinkov, Yonatan},
  journal={Computational Linguistics},
  volume={48},
  number={1},
  pages={207--219},
  year={2022}
}

@article{burns2022discovering,
  title={Discovering latent knowledge in language models without supervision},
  author={Burns, Collin and Ye, Haotian and Klein, Dan and Steinhardt, Jacob},
  journal={arXiv preprint arXiv:2212.03827},
  year={2022}
}

@inproceedings{azaria2023internal,
  title={The internal state of an LLM knows when it’s lying},
  author={Azaria, Amos and Mitchell, Tom},
  booktitle={Findings of the Association for Computational Linguistics: EMNLP 2023},
  pages={967--976},
  year={2023}
}

@article{marks2023geometry,
  title={The geometry of truth: Emergent linear structure in large language model representations of true/false datasets},
  author={Marks, Samuel and Tegmark, Max},
  journal={arXiv preprint arXiv:2310.06824},
  year={2023}
}

@article{zou2023representation,
  title={Representation engineering: A top-down approach to ai transparency},
  author={Zou, Andy and Phan, Long and Chen, Sarah and Campbell, James and Guo, Phillip and Ren, Richard and Pan, Alexander and Yin, Xuwang and Mazeika, Mantas and Dombrowski, Ann-Kathrin and others},
  journal={arXiv preprint arXiv:2310.01405},
  year={2023}
}

@article{li2023inference,
  title={Inference-time intervention: Eliciting truthful answers from a language model},
  author={Li, Kenneth and Patel, Oam and Vi{\'e}gas, Fernanda and Pfister, Hanspeter and Wattenberg, Martin},
  journal={Advances in neural information processing systems},
  volume={36},
  pages={41451--41530},
  year={2023}
}

@misc{cursor_python_sdk,
    author       = { Anysphere, Inc. },
    title        = { cursor-sdk: The official Python SDK for Cursor agent workflows },
    version = {1.0.29},
    year         = 2026,
    url          = { https://pypi.org/project/cursor-sdk/ },
    publisher    = { PyPI }
}

@misc{llama321b,
    author       = { Meta },
    title        = { Llama-3.2-1B  },
    year         = 2024,
    url          = { https://huggingface.co/meta-llama/Llama-3.2-1B },
    publisher    = { Hugging Face }
}

@misc{llama323b,
    author       = { Meta },
    title        = { Llama-3.2-3B  },
    year         = 2024,
    url          = { https://huggingface.co/meta-llama/Llama-3.2-3B},
    publisher    = { Hugging Face }
}

@misc{qwen2.50.5B,
    author       = { Qwen },
    title        = { Qwen2.5-0.5B-Instruct  },
    year         = 2024,
    url          = { https://huggingface.co/Qwen/Qwen2.5-0.5B-Instruct},
    publisher    = { Hugging Face }
}

@misc{qwen2.51.5B,
    author       = { Qwen },
    title        = { Qwen2.5-1.5B-Instruct  },
    year         = 2024,
    url          = { https://huggingface.co/Qwen/Qwen2.5-1.5B-Instruct},
    publisher    = { Hugging Face }
}

@misc{qwen2.57B,
    author       = { Qwen },
    title        = { Qwen2.5-7B-Instruct  },
    year         = 2024,
    url          = { https://huggingface.co/Qwen/Qwen2.5-7B-Instruct},
    publisher    = { Hugging Face }
}

@misc{qwen2.5,
    title = {Qwen2.5: A Party of Foundation Models},
    url = {https://qwenlm.github.io/blog/qwen2.5/},
    author = {Qwen Team},
    month = {September},
    year = {2024}
}

@article{qwen2,
      title={Qwen2 Technical Report}, 
      author={An Yang and Baosong Yang and Binyuan Hui and Bo Zheng and Bowen Yu and Chang Zhou and Chengpeng Li and Chengyuan Li and Dayiheng Liu and Fei Huang and Guanting Dong and Haoran Wei and Huan Lin and Jialong Tang and Jialin Wang and Jian Yang and Jianhong Tu and Jianwei Zhang and Jianxin Ma and Jin Xu and Jingren Zhou and Jinze Bai and Jinzheng He and Junyang Lin and Kai Dang and Keming Lu and Keqin Chen and Kexin Yang and Mei Li and Mingfeng Xue and Na Ni and Pei Zhang and Peng Wang and Ru Peng and Rui Men and Ruize Gao and Runji Lin and Shijie Wang and Shuai Bai and Sinan Tan and Tianhang Zhu and Tianhao Li and Tianyu Liu and Wenbin Ge and Xiaodong Deng and Xiaohuan Zhou and Xingzhang Ren and Xinyu Zhang and Xipin Wei and Xuancheng Ren and Yang Fan and Yang Yao and Yichang Zhang and Yu Wan and Yunfei Chu and Yuqiong Liu and Zeyu Cui and Zhenru Zhang and Zhihao Fan},
      journal={arXiv preprint arXiv:2407.10671},
      year={2024}
}

@misc{gemma31b,
    author       = { Google Deepmind },
    title        = { gemma-3-1b-it  },
    year         = 2025,
    url          = { https://huggingface.co/google/gemma-3-1b-it},
    publisher    = { Hugging Face }
}

@misc{gemma34b,
    author       = { Google Deepmind },
    title        = { gemma-3-4b-it },
    year         = 2025,
    url          = { https://huggingface.co/google/gemma-3-4b-it},
    publisher    = { Hugging Face }
}

@article{team2025gemma,
  title={Gemma 3 technical report},
  author={Team, Gemma and Kamath, Aishwarya and Ferret, Johan and Pathak, Shreya and Vieillard, Nino and Merhej, Ramona and Perrin, Sarah and Matejovicova, Tatiana and Ram{\'e}, Alexandre and Rivi{\`e}re, Morgane and others},
  journal={arXiv preprint arXiv:2503.19786},
  year={2025}
}
